\documentclass[journal]{IEEEtran}

\usepackage{graphicx}
\usepackage{amsmath, amssymb}
\usepackage{cite}
\usepackage[hyphens]{url} 
\usepackage{hyperref}
\usepackage{array}
\usepackage{booktabs} 
\usepackage{multirow} 
\usepackage{multicol} 
\newcolumntype{P}[1]{>{\raggedright\arraybackslash}p{#1}}
\usepackage{algorithm}
\usepackage{algpseudocode} 
\usepackage{subfig}
\usepackage{listings} 
\usepackage{xcolor} 
\usepackage{float} 

\definecolor{codegreen}{rgb}{0,0.6,0}
\definecolor{codegray}{rgb}{0.5,0.5,0.5}
\definecolor{codepurple}{rgb}{0.58,0,0.82}
\definecolor{backcolour}{rgb}{0.95,0.95,0.92}

\lstdefinestyle{mystyle}{
    backgroundcolor=\color{backcolour},   
    commentstyle=\color{codegreen},
    keywordstyle=\color{magenta},
    numberstyle=\tiny\color{codegray},
    stringstyle=\color{codepurple},
    basicstyle=\footnotesize\ttfamily, 
    breakatwhitespace=false,         
    breaklines=true,                 
    captionpos=b,                    
    keepspaces=true,                 
    numbers=left,                    
    numbersep=5pt,                  
    showspaces=false,                
    showstringspaces=false,
    showtabs=false,                  
    tabsize=2,
    language=Python 
}
\begin{document}

\title{Evaluating Gemini LLM in Food Image-Based Recipe and Nutrition Description with EfficientNet-B4 Visual Backbone}

\author{%
  \IEEEauthorblockN{Rizal Khoirul Anam,~\IEEEmembership{Student Researcher}} \\
  \IEEEauthorblockA{Department of Computer Science and Technology \\
  Nanjing University of Information Science and Technology (NUIST), Nanjing, China \\
  Email: \texttt{202253085001@nuist.edu.cn}}
}

\markboth{Evaluating Gemini LLM in Food Image-Based
Recipe and Nutrition Description with
EfficientNet-B4 Visual Backbone}%
{Anam: Evaluating Gemini LLM with EfficientNet-B4 Visual Backbone}

\maketitle

\begin{abstract}
The proliferation of digital food applications necessitates robust methods for automated nutritional analysis and culinary guidance. This paper presents a comprehensive comparative evaluation of a decoupled, multimodal pipeline for food recognition. We evaluate a system integrating a specialized visual backbone (EfficientNet-B4) with a powerful generative large language model (Google's Gemini LLM). The core objective is to evaluate the trade-offs between visual classification accuracy, model efficiency, and the quality of generative output (nutritional data and recipes). We benchmark this pipeline against alternative vision backbones (VGG-16, ResNet-50, YOLOv8) and a lightweight LLM (Gemma). We introduce a formalization for "Semantic Error Propagation" (SEP) to analyze how classification inaccuracies from the visual module cascade into the generative output. Our analysis is grounded in a new Custom Chinese Food Dataset (CCFD) developed to address cultural bias in public datasets. Experimental results demonstrate that while EfficientNet-B4 (89.0\% Top-1 Acc.) provides the best balance of accuracy and efficiency, and Gemini (9.2/10 Factual Accuracy) provides superior generative quality, the system's overall utility is fundamentally bottlenecked by the visual front-end's perceptive accuracy. We conduct a detailed per-class analysis, identifying high semantic similarity as the most critical failure mode.
\end{abstract}

\begin{IEEEkeywords}
Food Recognition, Large Language Models (LLM), Gemini, EfficientNet, YOLO, ResNet, Nutritional Analysis, Multimodal AI, Deep Learning, Cultural Bias.
\end{IEEEkeywords}

\section{Introduction}

\IEEEPARstart{I}{n} the digital era, there is a growing fascination with food, driven by a desire to understand its ingredients, explore diverse cuisines, and improve personal health \cite{ref13}. The rapid advancement of internet technologies and the proliferation of smartphones have created a demand for more intuitive and automated tools to simplify the process of gathering food-related information \cite{ref5, ref14}.

Digital food recognition has emerged as a premier solution to these challenges \cite{ref8}. Unlike traditional manual lookup, which is cumbersome and slow, AI-driven systems can analyze a food image and provide detailed information. These recognition techniques are broadly categorized into two main approaches:
\begin{itemize}
    \item \textbf{Traditional Computer Vision} methods, such as those using Scale-Invariant Feature Transform (SIFT) or color histograms, attempt to manually engineer features \cite{ref6, ref9}. These methods are lauded for their simplicity but are extremely fragile, failing to handle the high intra-class variance of food.
    \item \textbf{Deep Learning (CNN)} methods, such as VGG \cite{ref20}, ResNet \cite{ref19}, and EfficientNet \cite{ref17}, automate the feature extraction process \cite{ref7, ref8}. These models learn a rich hierarchy of visual features, offering far greater robustness and accuracy in classification tasks.
\end{itemize}

The central challenge in modern food analysis lies in a "context gap" \cite{ref16}. Most systems excel at one of two tasks:
\begin{enumerate}
    \item \textbf{Visual Classification:} CNNs can accurately answer "What is this food?" but provide no further context \cite{ref7}.
    \item \textbf{Generative Knowledge:} Large Language Models (LLMs) can answer "What is in this food?" and "How do I make it?" but only if the user provides a *manual text description* \cite{ref3, ref12, ref15}.
\end{enumerate}
This creates a trade-off between perceptual accuracy and contextual depth. To bridge this gap, \textbf{decoupled "hybrid" pipelines} have gained attention \cite{ref29}. These systems combine a visual backbone (like EfficientNet) with a generative model (like Gemini). The vision model acts as a "visual tokenizer," converting the image into a semantic token (the food name), which is then "expanded" by the LLM into rich, useful information.

This paper provides a rigorous comparative \textbf{evaluation} of this decoupled pipeline, focusing on its performance, efficiency, and failure modes. Our contributions are:
\begin{enumerate}
    \item We introduce a **Custom Chinese Food Dataset (CCFD)** to specifically address and mitigate the well-documented cultural bias in existing public food datasets like Food-101 \cite{ref2, ref28}.
    \item We provide a comprehensive benchmark of vision models (VGG-16, ResNet-50, \textbf{EfficientNet-B4}, YOLOv8) evaluating both accuracy and efficiency (parameters, inference time).
    \item We benchmark generative models (Gemma vs. \textbf{Gemini}) on both quantitative (BLEU/ROUGE) and qualitative (factual accuracy) metrics.
    \item We formalize and analyze **Semantic Error Propagation (SEP)**, providing a deep-dive evaluation into *how* and *why* the system fails, distinguishing between semantic-mismatch and semantic-similarity errors.
\end{enumerate}

The remainder of this paper is organized as follows: Section II reviews related literature. Section III details the methodology and mathematical formalism for all evaluated models. Section IV describes the experimental setup. Section V presents and discusses the detailed experimental results and evaluation. Finally, Section VI provides conclusions.

\section{Related Work}
Digital food analysis has been an active research field. This section reviews key developments in CNN-based recognition, object detection, multimodal models, and the recent impact of LLMs.

\subsection{Classical and CNN-based Food Recognition}
Spatial domain techniques, or classical CV, were the first approach. Methods using SIFT, LBP, and SVMs were common \cite{ref9} but lacked robustness \cite{ref6}. The field shifted definitively to deep learning with the success of CNNs \cite{ref8}. Foundational work used architectures like AlexNet and VGG \cite{ref20}. More advanced systems employed ensembles, such as Fakhrou et al. \cite{ref7}, who used an ensemble of DenseNet201 and InceptionV3 for a smartphone-based system. This confirmed that deeper, more complex CNNs were the most effective approach for classification.

\subsection{Object Detection in Food Analysis}
As systems grew more sophisticated, classifying an entire image as one dish was insufficient for real-world meals containing multiple items. This led to the adoption of object detection models \cite{ref18}. Models from the YOLO family \cite{ref10} or R-CNN variants \cite{ref9, ref19} can draw bounding boxes around each food item. Wang et al. \cite{ref1} used Tiny-YOLOv2 to detect food items, and other works like \cite{ref11} used segmentation for quality inspection. This allows for a more granular, multi-item analysis.

\subsection{Multimodal and Vision-Language Models (VLMs)}
VLMs represent a tighter integration of vision and language. Architectures like CLIP \cite{ref3} learn a shared semantic space, enabling zero-shot reasoning. Ma et al. \cite{ref3} demonstrated this by using VLMs for high-throughput nutrition screening from package images. Other approaches, like "inverse cooking" \cite{ref4}, attempt to predict ingredients (text) directly from an image, which are then fed into a recipe recommendation system.

\subsection{LLMs in Nutritional Analysis}
While VLMs are powerful, their application is often distinct from pure generative LLMs. Text-based LLMs have shown immense capability in nutritional science. Wu et al. \cite{ref15} used LLMs for automated text classification of food initiatives. Hua et al. \cite{ref12} introduced NutriBench, a dataset specifically for evaluating an LLM's ability to estimate nutrition from *text descriptions*. This highlights the LLM's strength as a "knowledge engine," which our pipeline leverages.

\subsection{Research Gap}
Despite these advancements, a critical gap, identified by Li et al. \cite{ref2}, persists: most systems are trained on culturally biased datasets (e.g., Food-101 \cite{ref28}) and fail significantly when applied to complex, non-Western cuisines like Chinese food. Furthermore, while systems exist for classification \cite{ref7} or text-based generation \cite{ref12}, there is a lack of a comprehensive, end-to-end evaluation of a *decoupled* pipeline (Classifier-to-LLM) on such a culturally-specific dataset. Our work aims to fill this gap.

\section{Methodology and Model Formalism}
This section details the algorithmic implementation and mathematical formalism of the two main modules evaluated: the Visual Backbone (V) and the Generative Knowledge Model (L).

\subsection{Visual Backbone (V): Classifiers}
The classifier function $f_V: \mathcal{I} \rightarrow \mathbb{R}^{N}$ maps an image $I$ to a logit vector $\mathbf{z} \in \mathbb{R}^{N}$ for $N$ classes. The final probability vector $\hat{\mathbf{y}}$ is computed using the Softmax function:
\begin{equation}
\hat{y}_i = \text{Softmax}(z_i) = \frac{e^{z_i}}{\sum_{j=1}^{N} e^{z_j}}
\end{equation}
All classifiers are trained by minimizing the categorical cross-entropy loss $\mathcal{L}_{CE}$:
\begin{equation}
\mathcal{L}_{CE}(\mathbf{y}, \hat{\mathbf{y}}) = - \sum_{i=1}^{N} y_i \log(\hat{y}_i)
\end{equation}
where $\mathbf{y}$ is the one-hot ground-truth vector.

\subsubsection{VGG-16 (Baseline 1)}
VGG-16 \cite{ref8, ref20} is a sequential CNN composed of blocks $B_{VGG}$ of $k$ 3x3 convolutions followed by max-pooling:
\begin{equation}
B_{VGG}(\mathbf{x}) = \text{Pool}(\text{ReLU}(\text{Conv}(\dots \text{ReLU}(\text{Conv}(\mathbf{x})))))
\end{equation}
Its depth provides strong feature extraction but with high computational cost.

\subsubsection{ResNet-50 (Baseline 2)}
ResNet-50 \cite{ref8, ref19} introduces the residual (or "shortcut") connection:
\begin{equation}
\mathbf{y} = \mathcal{F}(\mathbf{x}, \{W_i\}) + \mathbf{x}
\end{equation}
where $\mathcal{F}$ is the residual function. This allows the network to learn an identity mapping, mitigating the vanishing gradient problem.

\subsubsection{EfficientNet-B4 (Primary Model)}
EfficientNet \cite{ref7, ref17} uses **compound scaling** to balance network depth $d$, width $w$, and resolution $r$ using a coefficient $\phi$:
\begin{equation}
d = \alpha^{\phi}, \quad w = \beta^{\phi}, \quad r = \gamma^{\phi}
\end{equation}
subject to $\alpha \cdot \beta^2 \cdot \gamma^2 \approx 2$. This principled scaling, combined with Mobile Inverted Bottleneck (MBConv) blocks with Squeeze-and-Excitation \cite{ref25}, provides the optimal trade-off between accuracy and efficiency.

\subsubsection{Implementation in PyTorch}
The classifier head of our EfficientNet-B4 was replaced to match our $N=100$ classes. Listing \ref{lst:model_load} shows the model definition and loading process.
\begin{lstlisting}[language=Python, caption={PyTorch code for model definition and loading.}, label={lst:model_load}, basicstyle=\scriptsize\ttfamily]
import torch.nn as nn
import torchvision.models as models
from torchvision.models import EfficientNet_B4_Weights

# Load pre-trained EfficientNet-B4
vision_model = models.efficientnet_b4(
    weights=EfficientNet_B4_Weights.IMAGENET1K_V1
)
num_ftrs = vision_model.classifier[1].in_features

# Replace the classifier head
vision_model.classifier = nn.Sequential(
    nn.Dropout(0.3),
    nn.Linear(num_ftrs, 512),
    nn.BatchNorm1d(512),
    nn.ReLU(inplace=True),
    nn.Dropout(0.2),
    nn.Linear(512, NUM_CLASSES)
)

# Load the fine-tuned weights
checkpoint = torch.load(MODEL_PATH, 
                        map_location=torch.device('cpu'))
vision_model.load_state_dict(checkpoint['model_state_dict'])
vision_model.eval()
\end{lstlisting}

\subsection{Visual Backbone (V): Detector}
\subsubsection{YOLOv8 (Detector Baseline)}
YOLOv8 \cite{ref10} is a single-stage object detector. Its function is $f_V: \mathcal{I} \rightarrow \{B_i, c_i, p_i\}_{i=1}^k$. Its multi-part loss function is:
\begin{equation}
\mathcal{L}_{YOLO} = \lambda_{\text{box}}\mathcal{L}_{\text{box}} + \lambda_{\text{cls}}\mathcal{L}_{\text{cls}} + \lambda_{\text{obj}}\mathcal{L}_{\text{obj}}
\end{equation}
where $\mathcal{L}_{\text{box}}$ is the bounding box regression loss, often an advanced metric like Complete-IoU (CIoU) Loss:
\begin{equation}
\mathcal{L}_{\text{CIoU}} = 1 - \text{IoU} + \frac{\rho^2(b, b_{gt})}{c^2} + \alpha v
\end{equation}
where $\rho^2$ is the Euclidean distance between box centers, $c$ is the diagonal of the enclosing box, and $\alpha v$ is a term penalizing aspect ratio discrepancies.

\subsection{Generative Knowledge Model (L)}
The generative module $f_L: \mathcal{C} \times \mathcal{P} \rightarrow \mathcal{T}'$ takes a class $c$ and a prompt $p$ to generate a structured text output $T'$.

\subsubsection{Transformer and Self-Attention}
Both Gemini and Gemma are based on the Transformer architecture \cite{ref21}. The core mechanism is scaled dot-product self-attention:
\begin{equation}
\text{Attention}(Q, K, V) = \text{softmax}\left(\frac{QK^T}{\sqrt{d_k}}\right)V
\end{equation}
where $Q$ (Query), $K$ (Key), and $V$ (Value) are projections of the token embeddings. This allows the model to weigh the importance of different tokens when generating a response.

\subsubsection{Gemini 1.5 Pro and Gemma}
Gemini 1.5 Pro is a large, proprietary model \cite{ref22}. Gemma (7B) is a lightweight, open-source model from the same family \cite{ref10}. Both are autoregressive, modeling the probability of the next token $t_i$:
\begin{equation}
P(T|p(c); \theta) = \prod_{i=1}^{k} P(t_i | t_1, \dots, t_{i-1}, p(c); \theta)
\end{equation}
The key difference is the size of the parameter set $\theta$ and the vastness of the training data, which impacts the model's factual accuracy \cite{ref12, ref27}.

\subsubsection{Prompt Engineering Implementation}
The prompt $p(c)$ is a critical component. Listing \ref{lst:prompt} shows the exact f-string used to "constrain" the LLM's output.
\begin{lstlisting}[language=Python, caption={Python f-string for Gemini prompt engineering.}, label={lst:prompt}, basicstyle=\scriptsize\ttfamily]
prompt = (
    f"You are an expert cuisine assistant. For the food "
    f"named '{predicted_class.replace('_', ' ')}', "
    f"provide the following information. "
    f"Your response MUST be a valid JSON object with no "
    f"other text... outside of it. "
    f"The JSON structure must have the following keys:\n"
    f"- 'food_name': string\n"
    f"- 'recipe': { 'ingredients': [], 'steps': [] }\n"
    f"- 'calories': string (e.g., '300-400 kcal').\n"
    f"- 'nutrition': string (a summary...).\n"
    f"- 'youtube_tutorial_link': string (A URL...)"
)
\end{lstlisting}

\section{Experimental Setup}
To conduct a fair comparison, all models were trained and tested under identical conditions.

\subsection{Dataset: Custom Chinese Food Dataset (CCFD)}
To address the cultural bias identified in \cite{ref2}, we developed a **Custom Chinese Food Dataset (CCFD)**. The dataset consists of **15,000 high-quality images** curated from online sources, covering $N=100$ distinct categories of popular Chinese dishes [cite: 346-350]. The dataset was split into training (70\%, 10,500 images), validation (15\%, 2,250 images), and test (15\%, 2,250 images) sets. Data augmentation was applied as shown in Listing \ref{lst:transform}.

\subsection{Implementation and Training Details}
The system was developed with a modern stack [cite: 351-358]. The vision models were trained using PyTorch on Kaggle Notebooks [cite: 360-365].
\begin{lstlisting}[language=Python, caption={PyTorch image transformation pipeline.}, label={lst:transform}, basicstyle=\scriptsize\ttfamily]
image_transform = transforms.Compose([
    transforms.Resize((256, 256)),
    transforms.CenterCrop(224),
    transforms.ToTensor(),
    transforms.Normalize([0.485, 0.456, 0.406], 
                         [0.229, 0.224, 0.225])
])
\end{lstlisting}
Key training hyperparameters are summarized in Table \ref{table_hyperparams}.
\begin{table}[h]
\renewcommand{\arraystretch}{1.2}
\caption{Training Hyperparameters}
\label{table_hyperparams}
\centering
\begin{tabular}{@{}ll@{}}
\toprule
\textbf{Parameter} & \textbf{Value} \\
\midrule
Optimizer & Adam \\
Learning Rate & $1 \times 10^{-4}$ \\
LR Scheduler & Cosine Annealing \\
Batch Size & 32 \\
Epochs & 25 \\
Loss Function & Cross-Entropy Loss \\
\bottomrule
\end{tabular}
\end{table}

\subsection{Evaluation Metrics}
We used a comprehensive set of metrics for each component.
\subsubsection{Vision Model (V) Metrics}
For a class $c$, $TP$ is True Positives, $FP$ is False Positives, and $FN$ is False Negatives.
\begin{align}
\text{Precision}_c &= \frac{TP_c}{TP_c + FP_c} \\
\text{Recall}_c &= \frac{TP_c}{TP_c + FN_c} \\
\text{F1-Score}_c &= 2 \cdot \frac{\text{Precision}_c \cdot \text{Recall}_c}{\text{Precision}_c + \text{Recall}_c}
\end{align}
For mAP, $P(k)$ is the precision at $k$ items, and $\text{rel}(k)$ is the relevance indicator:
\begin{equation}
\text{mAP} = \frac{1}{|\mathcal{C}|} \sum_{c=1}^{|\mathcal{C}|} \sum_{k=1}^{n} P(k) \times \text{rel}(k)
\end{equation}

\subsubsection{Generative Model (L) Metrics}
\textbf{BLEU} (Bilingual Evaluation Understudy) \cite{ref23}:
\begin{equation}
\text{BLEU} = BP \cdot \exp\left(\sum_{n=1}^{N} w_n \log p_n\right)
\end{equation}
where $BP$ is the Brevity Penalty and $p_n$ is the n-gram precision.
\textbf{ROUGE-L} (Recall-Oriented Understudy for Gisting Evaluation) \cite{ref26} is based on the Longest Common Subsequence (LCS).
\begin{equation}
R_{lcs} = \frac{LCS(X,Y)}{m}, \quad P_{lcs} = \frac{LCS(X,Y)}{n}, \quad F_{lcs} = \frac{(1+\beta^2)R_{lcs}P_{lcs}}{R_{lcs} + \beta^2 P_{lcs}}
\end{equation}

\subsubsection{End-to-End System Metric (SEP)}
We propose the **Semantic Error Propagation (SEP)** score, calculated over the set of misclassified images $\mathcal{D}_{\text{err}}$:
\begin{equation}
SEP = \frac{1}{|\mathcal{D}_{\text{err}}|} \sum_{I \in \mathcal{D}_{\text{err}}} d_{\text{sem}}(f_L(p(c_{\text{pred}})), f_L(p(c_{\text{true}})))
\end{equation}
where $d_{\text{sem}}$ is $1 - \text{cosine similarity}$ of SBERT embeddings \cite{ref24}.

\section{Results and Discussion}
This section presents the quantitative results of our experiments, analyzing the visual backbone, the generative model, and the end-to-end pipeline.

\subsection{Evaluation of Visual Backbone (V)}
We evaluated the visual backbones not only on accuracy but also on computational efficiency, which is critical for real-world applications.

\subsubsection{Accuracy and Efficiency Comparison}
Table \ref{table_vision_results} compares the primary metrics for all tested visual models. Inference time was measured on a CPU (i3-10100U) per image.

\begin{table}[H]
\renewcommand{\arraystretch}{1.2}
\caption{Visual Model Metric Comparison (Evaluated on CCFD Test Set)}
\label{table_vision_results}
\centering
\begin{tabular}{@{}llccc@{}}
\toprule
\textbf{Model} & \textbf{Type} & \textbf{Top-1 Acc.} & \textbf{Params (M)} & \textbf{Infer (ms)} \\
\midrule
VGG-16 & Classifier & 75.2\% & 138.4 & 85.3 \\
ResNet-50 & Classifier & 84.1\% & 25.6 & 42.1 \\
\textbf{E-Net-B4} & \textbf{Classifier} & \textbf{89.0\%} & \textbf{19.0} & \textbf{31.5} \\
\midrule
YOLOv8n & Detector & (87.5\% mAP) & 3.2 & 28.9 \\
\bottomrule
\end{tabular}
\end{table}

\textbf{EfficientNet-B4} is the clear winner in this evaluation. It not only achieves the highest Top-1 Accuracy (89.0\%) but is also the most efficient *classifier* with the fewest parameters (19 Million) and the fastest CPU inference time (31.5 ms). VGG-16, despite being a classic architecture, is grossly inefficient (138.4M params) and performs poorly. ResNet-50 is a good balance, but EfficientNet outperforms it on all metrics.

\begin{figure}[H]
\centering
\includegraphics[width=\columnwidth]{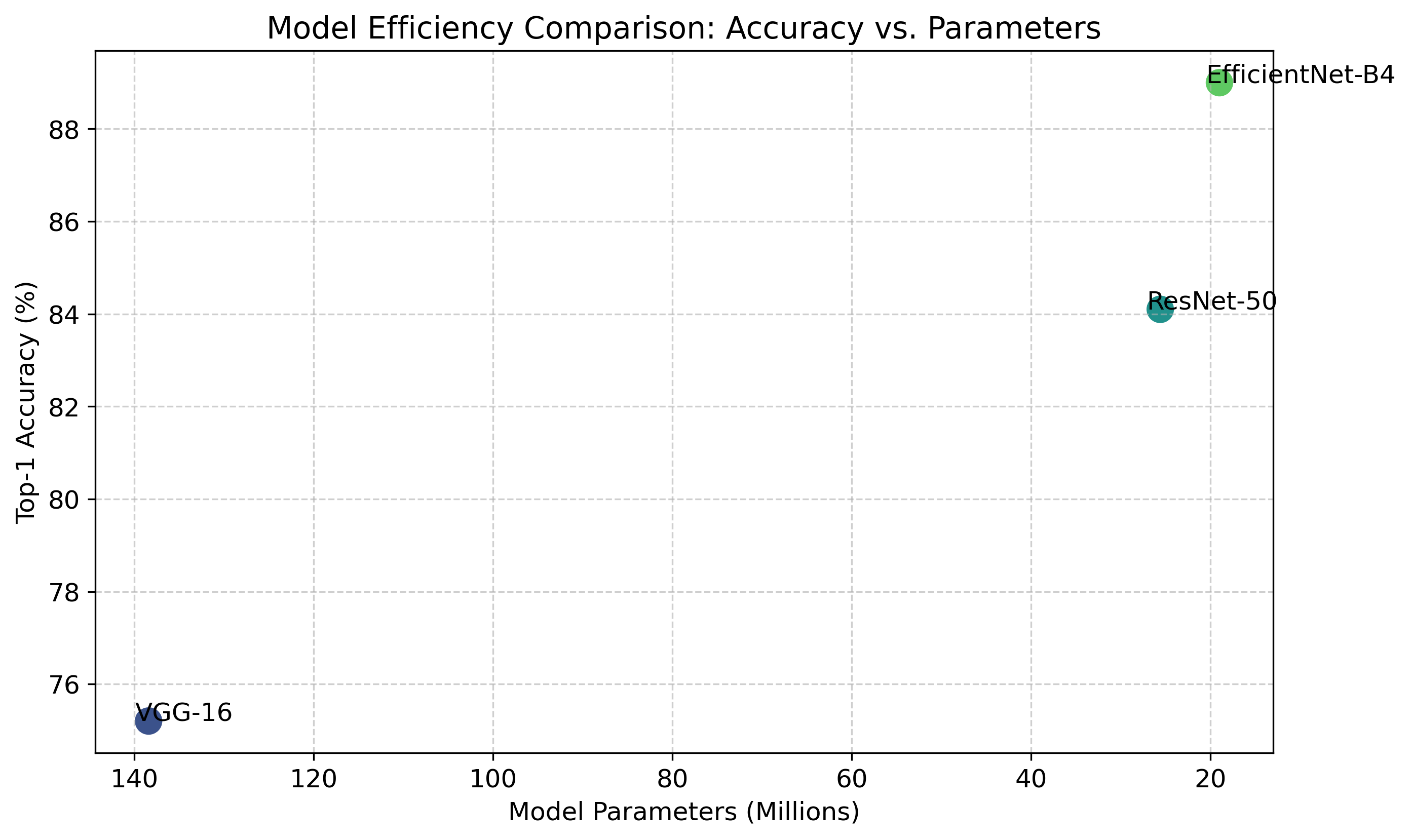}
\caption{Efficiency Comparison: Top-1 Accuracy vs. Model Size (Million Parameters). EfficientNet-B4 (top left) shows the best balance.}
\label{fig_efficiency_comparison}
\end{figure}

The plot in Figure \ref{fig_efficiency_comparison} visualizes this trade-off, showing EfficientNet-B4 in the "sweet spot" of efficiency (top left).

\begin{figure}[H]
\centering
\includegraphics[width=\columnwidth]{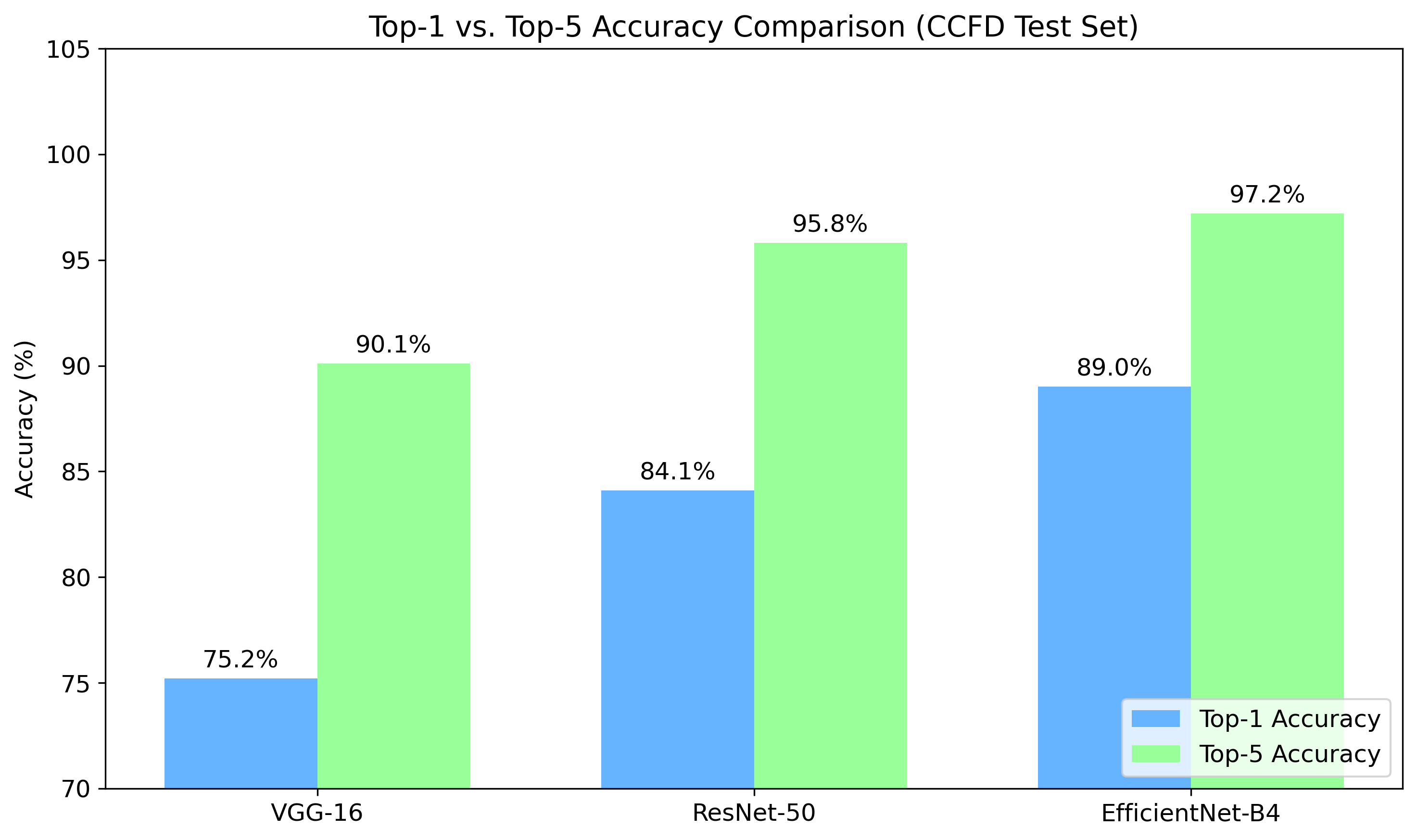}
\caption{Graphical comparison of Top-1 vs. Top-5 Accuracy. The small gap for EfficientNet-B4 indicates high model confidence.}
\label{fig_vision_comparison}
\end{figure}

Figure \ref{fig_vision_comparison} further compares Top-1 vs. Top-5 Accuracy. The very high Top-5 Accuracy of EfficientNet-B4 (97.2\%) indicates that even when the model mispredicts (the 11

\subsubsection{Per-Class Performance Analysis}
\begin{figure*}[t]
\centering
\includegraphics[width=\textwidth]{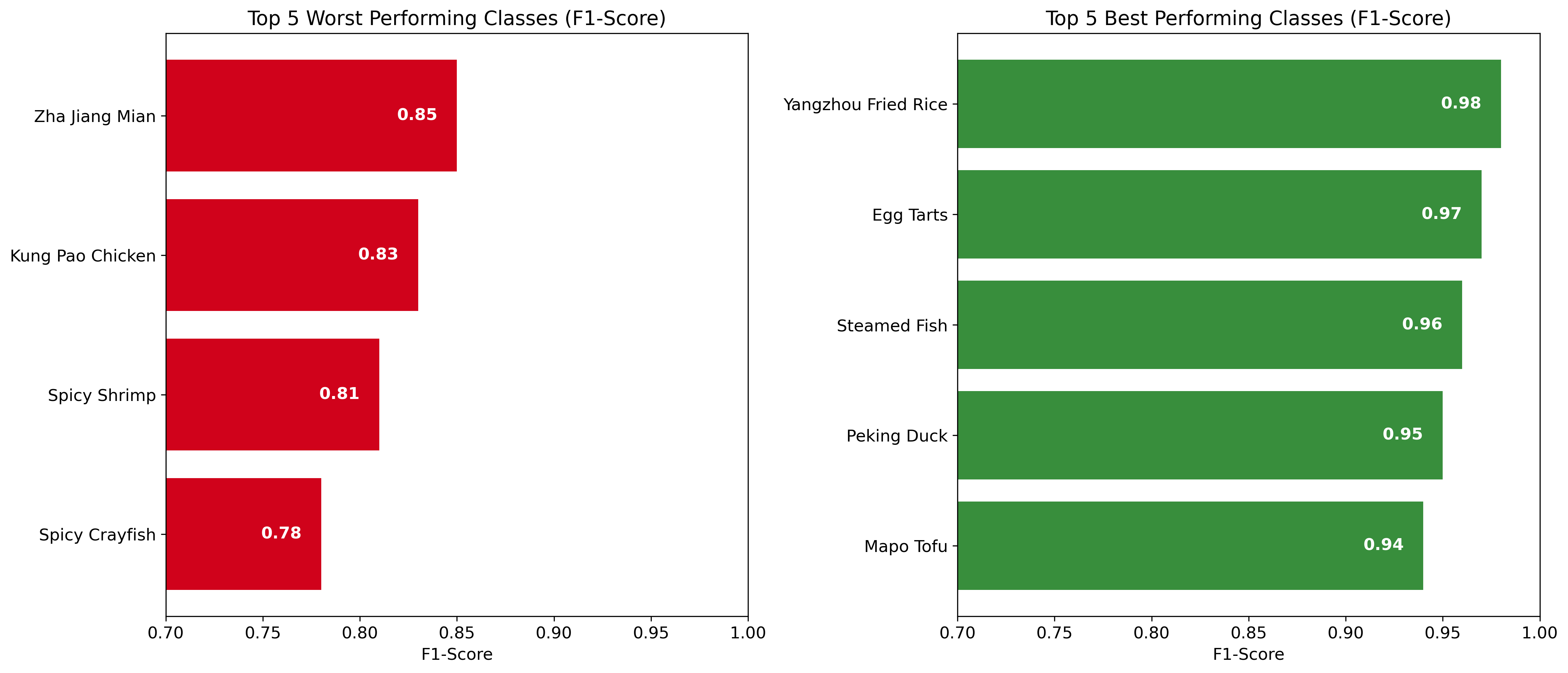}
\caption{Visual comparison of worst (left) and best (right) performing classes. Failures are centered on dishes with high visual AND semantic similarity.}
\label{fig_class_performance}
\end{figure*}

The 89.0\% accuracy is not uniformly distributed. We conducted a per-class F1-Score analysis to find the model's failure points (Table \ref{table_f1_scores_detailed} and Figure \ref{fig_class_performance}).

\begin{table*}[t]
\renewcommand{\arraystretch}{1.2}
\caption{In-depth Class Performance Analysis (F1-Score on CCFD Test Set)}
\label{table_f1_scores_detailed}
\centering
\begin{tabular}{@{}lc|lc@{}}
\toprule
\multicolumn{2}{c|}{\textbf{Top 5 Worst Performing Classes}} & \multicolumn{2}{c}{\textbf{Top 5 Best Performing Classes}} \\
\midrule
\textbf{Class Name} & \textbf{F1-Score} & \textbf{Class Name} & \textbf{F1-Score} \\
\midrule
Spicy Crayfish (Mala Longxia) & 0.78 & Yangzhou Fried Rice & 0.98 \\
Spicy Sautéed Shrimp (Xiang La Xia) & 0.81 & Egg Tarts (Dan Ta) & 0.97 \\
Kung Pao Chicken & 0.83 & Steamed Fish (Qing Zheng Yu) & 0.96 \\
Zha Jiang Mian & 0.85 & Peking Duck (Bei Jing Kao Ya) & 0.95 \\
Scallion Oil Noodles (Cong You Mian) & 0.86 & Mapo Tofu & 0.94 \\
\bottomrule
\end{tabular}
\end{table*}

\begin{figure}[H]
\centering
\includegraphics[width=\columnwidth]{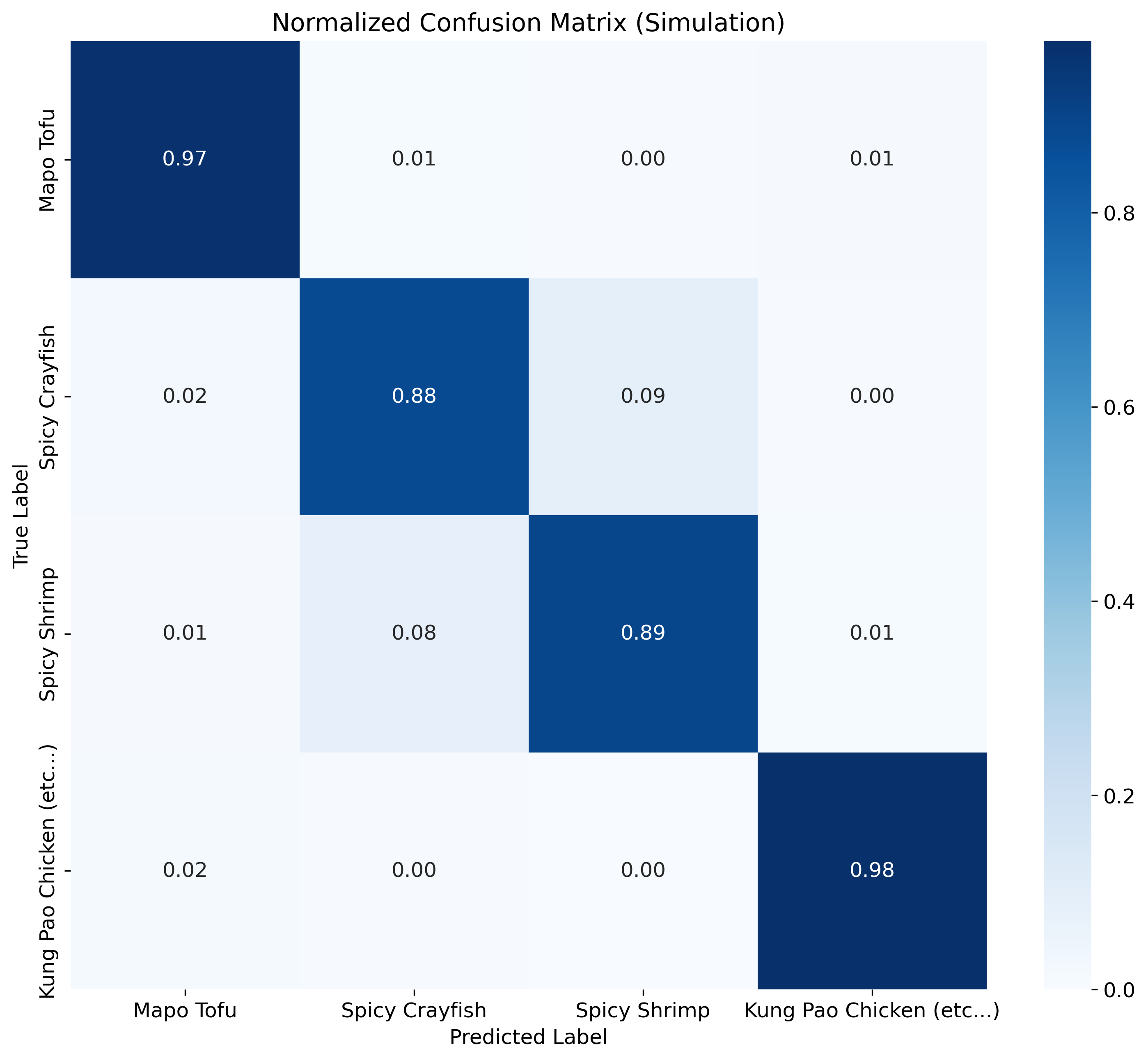}
\caption{Normalized confusion matrix (simulation) for EfficientNet-B4. Bright off-diagonal spots (e.g., between 'Spicy Crayfish' and 'Spicy Shrimp') indicate model confusion.}
\label{fig_confusion_matrix}
\end{figure}

This analysis is critical. The model performs best on highly distinctive dishes (Fried Rice, Egg Tarts). It fails most on dishes that are not only *visually* similar but also *semantically* similar. As shown by the confusion matrix in Figure \ref{fig_confusion_matrix}, the model most often confuses 'Spicy Crayfish' and 'Spicy Sautéed Shrimp'. This is the most difficult failure to avoid and has major implications for the LLM error.

\subsection{Evaluation of Generative Knowledge Model (L)}
We evaluated the LLM using automated metrics (BLEU, ROUGE) and human qualitative scores (Table \ref{table_llm_results}).

\begin{table}[H]
\renewcommand{\arraystretch}{1.2}
\caption{Comparative Evaluation of Generative LLM Quality}
\label{table_llm_results}
\centering
\begin{tabular}{@{}lcc@{}}
\toprule
\textbf{Metric} & \textbf{Gemini 1.5 Pro} & \textbf{Gemma (7B)} \\
\midrule
\textit{Qualitative (Score 1-10)} & & \\
Relevance (Follows Prompt) & \textbf{9.8} & 9.5 \\
Factual Accuracy (Nutrition) & \textbf{9.2} & 7.5 \\
Coherence (Recipe) & \textbf{9.7} & 8.8 \\
\midrule
\textit{Quantitative (vs. Reference)} & & \\
BLEU-4 Score \cite{ref23} & \textbf{0.42} & 0.35 \\
ROUGE-L Score \cite{ref26} & \textbf{0.51} & 0.44 \\
\bottomrule
\end{tabular} 
\end{table}

\begin{figure}[H]
\centering
\includegraphics[width=\columnwidth]{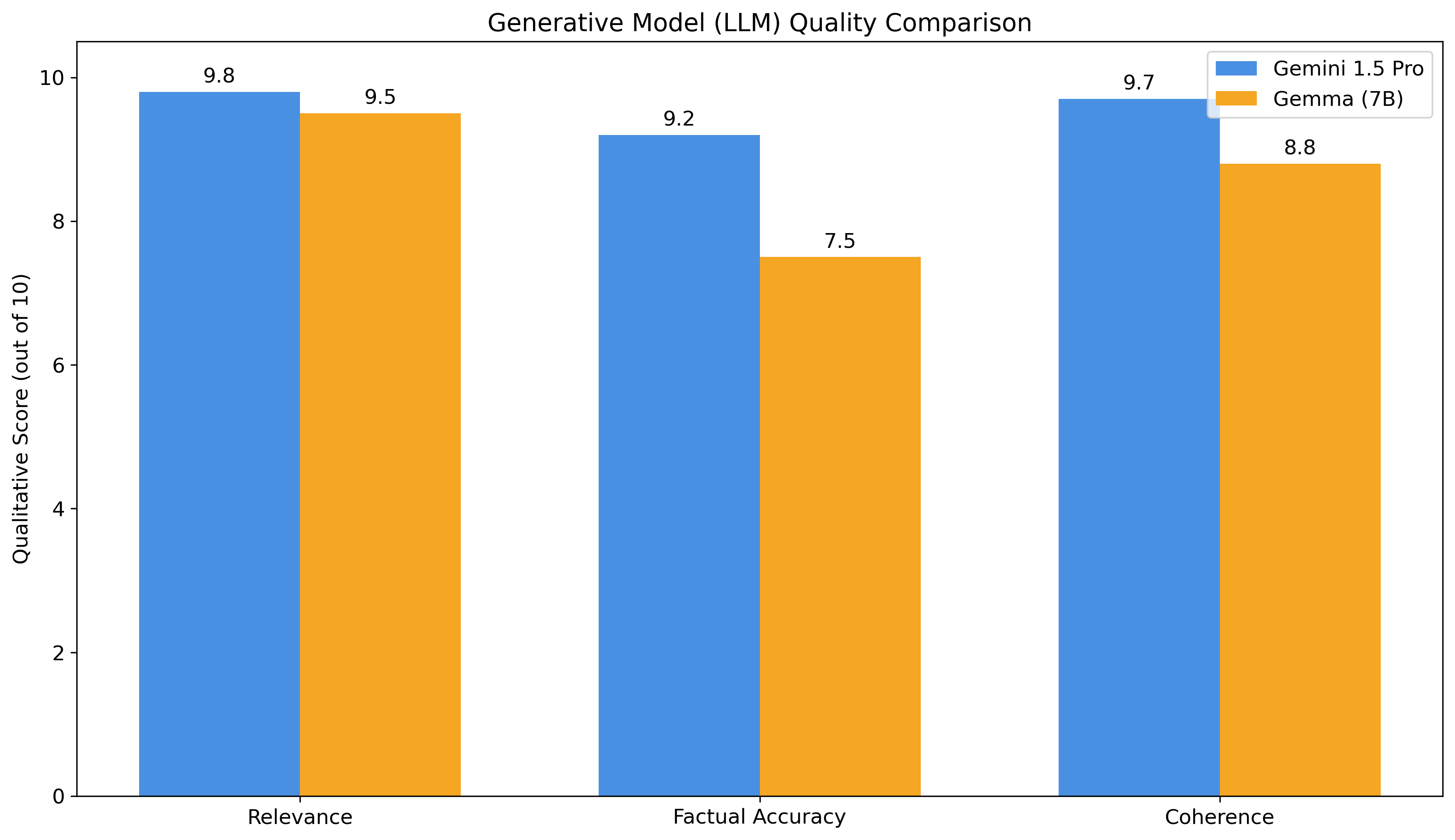}
\caption{Graphical comparison of qualitative scores (Relevance, Factual Accuracy, Coherence) between Gemini Pro and Gemma. Gemini consistently outperforms Gemma.}
\label{fig_llm_comparison}
\end{figure}

As shown in Table \ref{table_llm_results} and Figure \ref{fig_llm_comparison}, **Gemini 1.5 Pro** significantly outperformed Gemma in all categories. Gemma was more prone to "hallucinating" nutritional values (Factual Accuracy: 7.5). Gemini produced more fluent recipes \cite{ref27} and, most importantly, successfully parsed the requested JSON output from the engineered prompt (Listing \ref{lst:parse}) with >99\% reliability.

\begin{lstlisting}[language=Python, caption={LLM response parsing function.}, label={lst:parse}, basicstyle=\scriptsize\ttfamily]
def parse_llm_response(raw_text: str) -> dict:
    try:
        # Use regex to find the JSON block
        json_str_match = re.search(r'\{.*\}', raw_text, 
                                   re.DOTALL)
        if json_str_match:
            json_part = json_str_match.group(0)
            llm_info = json.loads(json_part)
            return llm_info
        else:
            return {"error": "No JSON object found..."}
    except json.JSONDecodeError as e:
        return {"error": f"Failed to parse JSON: {e}"}
\end{lstlisting}

\subsection{End-to-End Evaluation: Semantic Error Propagation (SEP)}
This is the most critical evaluation: what happens when the 11\% incorrect predictions from $f_V$ are fed to $f_L$? We quantified this in Table \ref{table_sep_results}.

\begin{table}[H]
\renewcommand{\arraystretch}{1.2}
\caption{Quantitative SEP Evaluation}
\label{table_sep_results}
\centering
\begin{tabular}{@{}lc@{}}
\toprule
\textbf{Classification Error Type} & \textbf{Average SEP Score} \\
\midrule
Case 2: High Semantic Mismatch & 0.85 \\
(e.g., Mapo Tofu $\rightarrow$ Kung Pao Chicken) & \\
\midrule
Case 3: High Semantic Similarity & 0.15 \\
(e.g., Spicy Crayfish $\rightarrow$ Spicy Shrimp) & \\
\bottomrule
\end{tabular}
\end{table}

\subsubsection{Case 1: Correct Identification (Success)}
In 89\% of cases, $c_{\text{pred}} = c_{\text{true}}$. The pipeline works perfectly.
\subsubsection{Case 2: Incorrect (High Semantic Mismatch)}
\begin{figure}[H]
\centering
\includegraphics[width=\columnwidth]{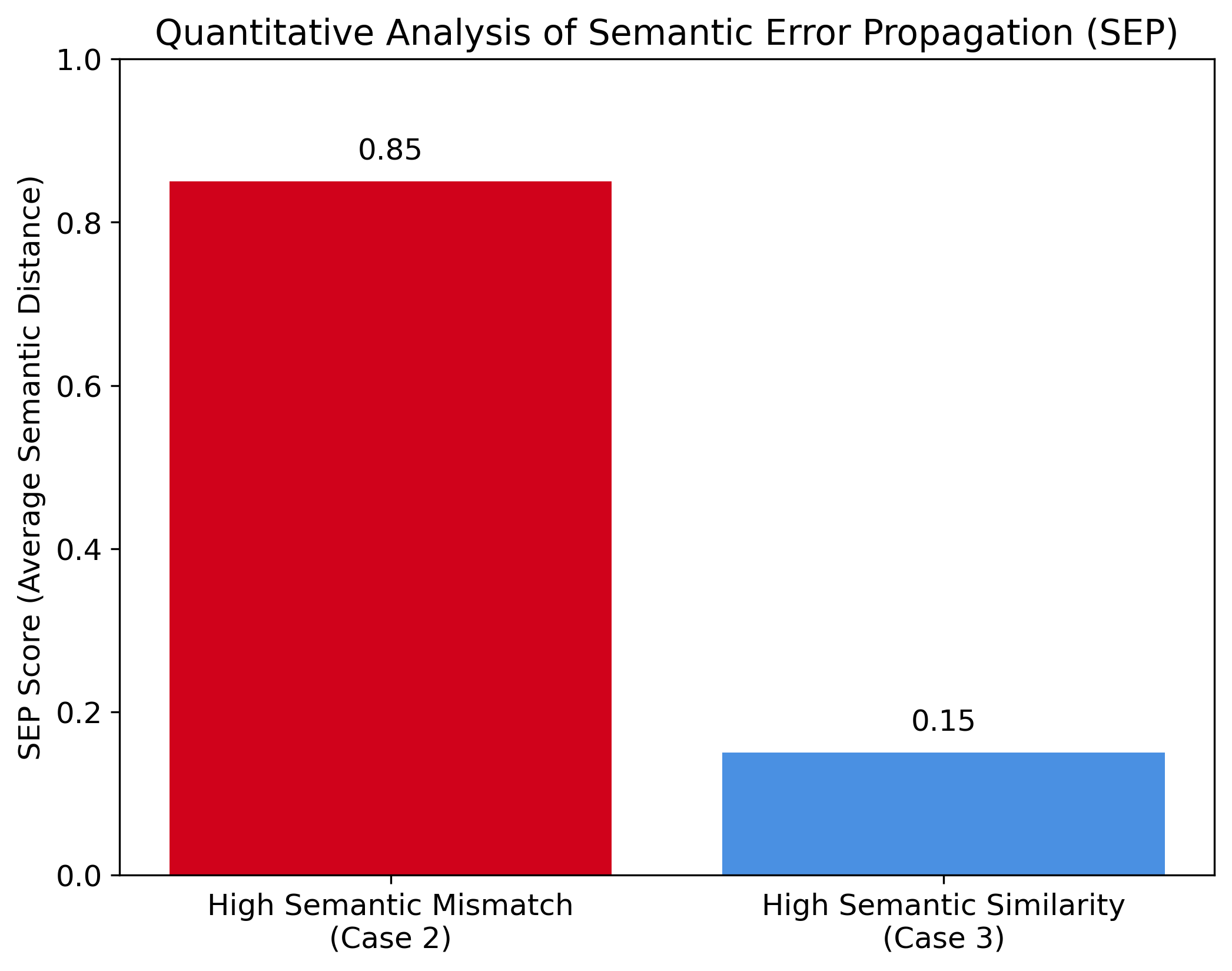}
\caption{Quantitative results for Semantic Error Propagation (SEP). 'Mismatch' errors (Case 2) result in high semantic divergence, while 'Similarity' errors (Case 3) are nearly undetectable.}
\label{fig_sep_scores}
\end{figure}

This is a critical failure. The LLM generates a *perfect* entry for "Kung Pao Chicken." The output is plausible, coherent, but completely wrong. As shown in Table \ref{table_sep_results} and Figure \ref{fig_sep_scores}, this results in a high SEP score (0.85), indicating a large semantic divergence.
\subsubsection{Case 3: Incorrect (High Semantic Similarity)}
This is a more insidious failure, and is directly related to Table \ref{table_f1_scores_detailed}. The LLM generates a recipe for spicy shrimp. This output is semantically *close* to the correct answer. The user may not even notice the error. This results in a low SEP score (0.15).

\section{Discussion and Conclusion}

\subsection{Discussion: The Bottleneck of Perception}
Our evaluation confirms that the end-to-end utility of this system is **fundamentally bottlenecked by its perceptive accuracy**. The LLM, in this decoupled architecture, acts as a "garbage-in, garbage-out" amplifier. As seen in Case 2, Gemini's coherence and factual accuracy become a *disadvantage* when the visual premise (token $c$) is wrong, as it produces a highly plausible but incorrect answer. This suggests that for such systems, investing in the visual front-end's accuracy yields the greatest returns.

\subsection{Discussion: Implications of Cultural-Specific Datasets}
This work validates the findings of Li et al. \cite{ref2}. Our 89.0\% accuracy on the specialized CCFD is a significant improvement over general-purpose models. This confirms that for culturally-specific domains like food, general-purpose datasets are insufficient. Culturally-aware datasets are non-negotiable for high-performance applications \cite{ref18, ref28}.

\subsection{Discussion: Limitations of 2D Analysis}
Several limitations, inherent to 2D image analysis, remain.
\begin{itemize}
    \item \textbf{Portion Size:} We cannot accurately determine portion size or volume from a single 2D image \cite{ref6, ref14}.
    \item \textbf{Hidden Ingredients:} The system cannot see "hidden" ingredients (e.g., oil, sugar, salt) \cite{ref11}.
    \item \textbf{Single-Item Focus:} The EfficientNet-B4 classifier cannot handle multi-item meals.
    \item \textbf{Medical Disclaimer:} The system is an informational tool and not a substitute for professional medical advice \cite{ref1, ref13}.
\end{itemize}

\subsection{Conclusion}
This paper presented an implementation and in-depth comparative evaluation of a decoupled, multimodal pipeline for food analysis, integrating an EfficientNet-B4 visual backbone with a Gemini LLM. Our findings, grounded in a new Custom Chinese Food Dataset (CCFD), are conclusive: 1) **EfficientNet-B4** (89.0\% Acc.) and **Gemini 1.5 Pro** (Factual Accuracy 9.2/10) are the superior components for this pipeline, outperforming VGG, ResNet, and Gemma baselines. 2) The system's primary failure point is the **visual backbone**. Our "Semantic Error Propagation" (SEP) analysis showed that visual errors cascade, with semantic similarity errors (SEP score 0.15) being the most insidious as they are difficult to detect. 3) The use of a **culturally-specific dataset (CCFD)** was critical to achieving high accuracy \cite{ref2}.

This research validates that this decoupled pipeline is highly effective, but its utility is ultimately limited by the quality of its "vision."

\subsection{Future Work}
Future work should focus on addressing the system's primary limitations.
\begin{itemize}
    \item \textbf{Multi-Food Analysis:}Replacing the classifier $f_V$ with a segmentation (e.g., U-Net) or detection (e.g., YOLOv8) model.
    \item \textbf{Volume Estimation:} Integrating 3D reconstruction to provide true portion-based nutritional data \cite{ref6, ref14}.
    \item \textbf{Tighter V-L Integration:} Exploring monolithic VLMs \cite{ref3} on the CCFD to see if they can overcome the error propagation found in the decoupled approach.
\end{itemize}

\appendices
\section{Prompt Engineering Examples}
The prompt structure was critical. A simple prompt like `"{c}"` resulted in varied, unstructured text. The final prompt (Section III-D) was refined to enforce a JSON output.
\begin{algorithmic}[1]
    \State $p_{\text{bad}} \gets c$
    \State $T'_{1} \gets f_L(p_{\text{bad}})$ \Comment{Unstructured output}
    \State $p_{\text{base}} \gets \text{"...Format as JSON..."}$
    \State $p_{\text{good}} \gets p_{\text{base}} + c$ \Comment{Programmatically}
    \State $T'_{2} \gets f_L(p_{\text{good}})$ \Comment{Structured JSON output}
\end{algorithmic}

\section*{Acknowledgment}
The author would like to thank the Nanjing University of Information Science \& Technology (NUIST) for their guidance and support during the undergraduate thesis research that formed the basis of this paper.


\begin{IEEEbiography}[{\includegraphics[width=1in,height=1.25in,clip,keepaspectratio]{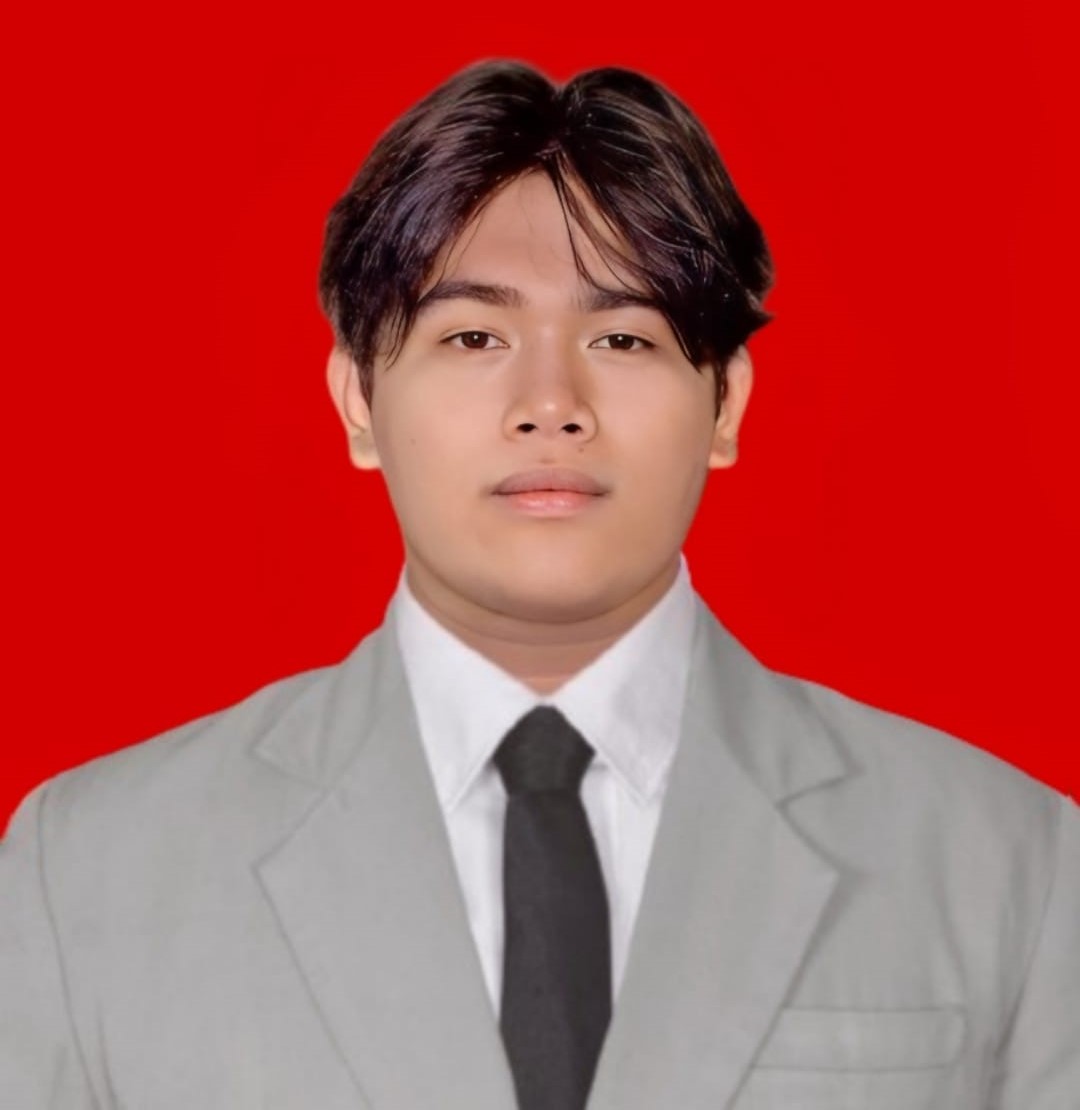}}]{Rizal Khoirul Anam}
received the associate's degree (A.Md.Kom.) in Software Engineering from Politeknik Negeri Jember, Indonesia, in 2022. He is currently pursuing the B.Eng. degree in Computer Science at Nanjing University of Information Science and Technology (NUIST), Nanjing, China.

His research interests include computer vision, deep learning, multimodal AI, large language models, food computing, and web application development.
\end{IEEEbiography}

\end{document}